\documentclass[sigconf]{acmart}

\setcopyright{none}
\renewcommand\footnotetextcopyrightpermission[1]{} 

\usepackage{amsmath,amsfonts}
\usepackage{algorithmic}
\usepackage{graphicx}
\usepackage{textcomp}
\usepackage{xcolor}
\usepackage{epsfig,caption,url,booktabs, subcaption}
\usepackage[nameinlink,capitalize]{cleveref}
\usepackage{tikz}

\usetikzlibrary{arrows.meta}

\DeclareMathOperator*{\argmax}{arg\,max}

\AtBeginDocument{%
  \providecommand\BibTeX{{%
    \normalfont B\kern-0.5em{\scshape i\kern-0.25em b}\kern-0.8em\TeX}}}

\setcopyright{acmcopyright}
\copyrightyear{2021}
\acmYear{2021}

\acmConference[SIGMOD '22]{SIGMOD '22}{June 12--17, 2022}{Philadelphia, PA, USA}

\begin{document}

\title{RLFlow: Optimising Neural Network Subgraph Transformation with World Models}





\author{Sean Parker}
\affiliation{%
  \institution{University of Cambridge}
  \city{Cambridge}
  \country{UK}
}
\email{sjp240@cantab.ac.uk}

\author{Sami Alabed}
\affiliation{%
  \institution{University of Cambridge}
  \city{Cambridge}
  \country{UK}
}
\email{sa894@cam.ac.uk}

\author{Eiko Yoneki}
\affiliation{%
  \institution{University of Cambridge}
  \city{Cambridge}
  \country{UK}
}
\email{eiko.yoneki@cl.cam.ac.uk}

\begin{abstract}
Training deep learning models takes an extremely long execution time and consumes large amounts of computing resources. At the same time, recent research proposed systems and compilers that are expected to decrease deep learning models runtime. An effective optimisation methodology in data processing is desirable, and the reduction of compute requirements of deep learning models is the focus of extensive research.

In this paper, we address the neural network sub-graph transformation by exploring reinforcement learning (RL) agents to achieve performance improvement. Our proposed approach \textbf{RLFlow} can learn to perform neural network subgraph transformations, without the need for expertly designed heuristics to achieve a high level of performance.

Recent work has aimed at applying RL to computer systems with some success, especially using model-free RL techniques. Model-based reinforcement learning methods have seen an increased focus in research as they can be used to learn the transition dynamics of the environment; this can be leveraged to train an agent using a hallucinogenic environment such as World Model (WM) \cite{ha2018worldmodels}, thereby increasing sample efficiency compared to model-free approaches. WM uses variational auto-encoders and it builds a model of the system and allows exploring the model in an inexpensive way.

In \textbf{RLFlow}, we propose a design for a model-based agent with WM which learns to optimise the architecture of neural networks by performing a sequence of sub-graph transformations to reduce model runtime. We show that our approach can match the state-of-the-art performance on common convolutional networks and outperforms by up to 5\% those based on transformer-style architectures
\end{abstract}



\maketitle

\pagestyle{plain}

\section{Introduction}
Recent modern software has key components that are underpinned by machine learning (ML) models, specifically, deep neural networks (DNN). Over the past decade, there has been a focus on developing frameworks that provide tools using which we can design, train and evaluate these deep learning models.

A common internal representation for neural networks inside deep learning frameworks is that of a computation graph; a directed acyclic graph where nodes represent a specific computation and edges the paths where data is transferred. It is a common optimisation practice to support graph substitutions. Frameworks such as TensorFlow \cite{tensorflow2015whitepaper, 199317} and PyTorch \cite{paszke2019pytorch} automatically apply optimisations in an effort to reduce computation resources during inference.

Currently, the majority of optimisations in deep learning frameworks are performed using manually defined heuristics. TensorFlow \cite{tensorflow2015whitepaper, 199317}, TensorRT \cite{tensorrt2017}, and TVM \cite{chen2018tvm} perform substitutions to a computation graph by using rule-based strategies. For example, TensorFlow \cite{tensorflow2015whitepaper, 199317} uses 155 handwritten optimisations composed of 53,000 lines of C++. While such heuristics are applicable for current architectures, network design is consistently evolving. Therefore, we require consistent innovation to discover and design rules that control the application of optimisations with guarantees that strictly improve efficiency. Such an approach is difficult to scale. Eliminating the need for manual engineering work that is required to design and implement the heuristics for applying optimisations is a primary focus of our work.

Recent work, namely TASO \cite{jia2019taso}, has shown the substitution of the subgraphs by replacing heuristics with an automatic cost-based search. However, such approaches may not fully explore the potential search space due to the lack of forward planning in cost-based optimisation. As a step towards resolving the issue of poor exploration, this work explores the use of reinforcement learning (RL). RL is an area of machine learning in which an agent learns to act optimally, given a state and a suitable reward function, through interactions with an environment.

In this paper, we introduce \textbf{RLFlow}, a deep learning graph optimiser that uses Reinforcement Learning for automating graph substitutions. Specifically, we focus on model-based reinforcement learning that aims to learn a model of the environment in which the agents act. We use an approach called ``World Models'' (WM), first proposed by Ha et al. \cite{ha2018worldmodels}. WM use variational auto-encoders, where at first randomly collected episode rollouts are used as training data to build a compact model, followed by training the compact model with Recurrent Neural Networks (RNN) to predict future steps to maximise the expected cumulative reward of rollouts. When using a WM as a simulated environment, several model evaluations can occur safely in parallel and, especially in systems environments, it overcomes the latency impact of updating system environments.

In this work, the network learns to model the dynamics of sub-graph transformations in a deep learning model as well as the impact on the overall runtime of the model when executed on-device. Further, learning a model of the environment provides important benefits; for example, lookahead planning, low-cost state prediction and faster wall-clock training. We examine the use of world models for learning the environment as well as training a controller inside a world model which removes the need for an expensive, time-consuming computer system to apply the chosen subgraph transformations.

Note that training and inferring on Sequence-to-Sequence (Seq2Seq) \cite{sutskever2014sequence} models differ from the usual classification problem, which characterises also for Transformers \cite{vaswani2017attention}. One of the focuses of our work in this paper is the applicability of RL based graph transformation over such transformer type of models. Transformers are getting very popular beyond Natural Language Processing (NLP) and it will inevitably expand further.

RLFlow is a graph-net based neural network optimisation as an extension of TASO. The basic idea was to substitute TASO’s cost-based backtracking search and replace it with an RL-based optimisation. This enables it to generalise to large unseen graphs and thus, find better-performing solutions than a backtracking search.

To summarise, our contributions are:

\begin{itemize}

\item We designed a model-based RL agent and environment for jointly choosing the optimal substitution and substitution location. Our model-based reinforcement learning approach eliminates the need for human-engineered graph optimisations in machine learning frameworks. We show that our approach can improve model runtime by up to 58\% compared to current deep learning frameworks.

\item We find a novel transformation of popular DNN architectures previously were unfeasible, where the agent isn't getting stuck in the greedy cost-based exploration, but rather is able to actually learn what is a good transformation given a context.

\item Our approach is especially powerful for transformer-style architectures demonstrating that it outperforms state-of-the-art methods by up to 5\%.

\item We provide a detailed discussion and analysis of our solution as well as comparison to the state-of-the-art methods in published literature.

\item This work, to the best of our knowledge, is the first that has applied model-based reinforcement learning in optimising computation graphs.

\end{itemize}

\section{Background}

\subsection{Computation graphs for neural networks}

There has been a rapid development of various deep learning architectures to solve specific groups of tasks. Common examples include convolutional networks, for a variety of tasks such as object detection and classification. Transformer networks for translation and generation of language. Finally, Recurrent networks have been shown to excel at exploiting long and short term trends in data.

Despite the improvements in the accuracy of machine learning (ML) models, the fundamental building blocks of deep learning models have remained largely unchanged. As the networks become more complex, it also becomes tedious to manually optimise the networks to reduce the execution time on hardware. Therefore, there is extensive work to automatically optimise the models, or alternatively, apply a set of hand-crafted optimisations according to pre-defined rules.

\begin{figure}[ht]
  \centering
  \includegraphics[width=0.75\columnwidth]{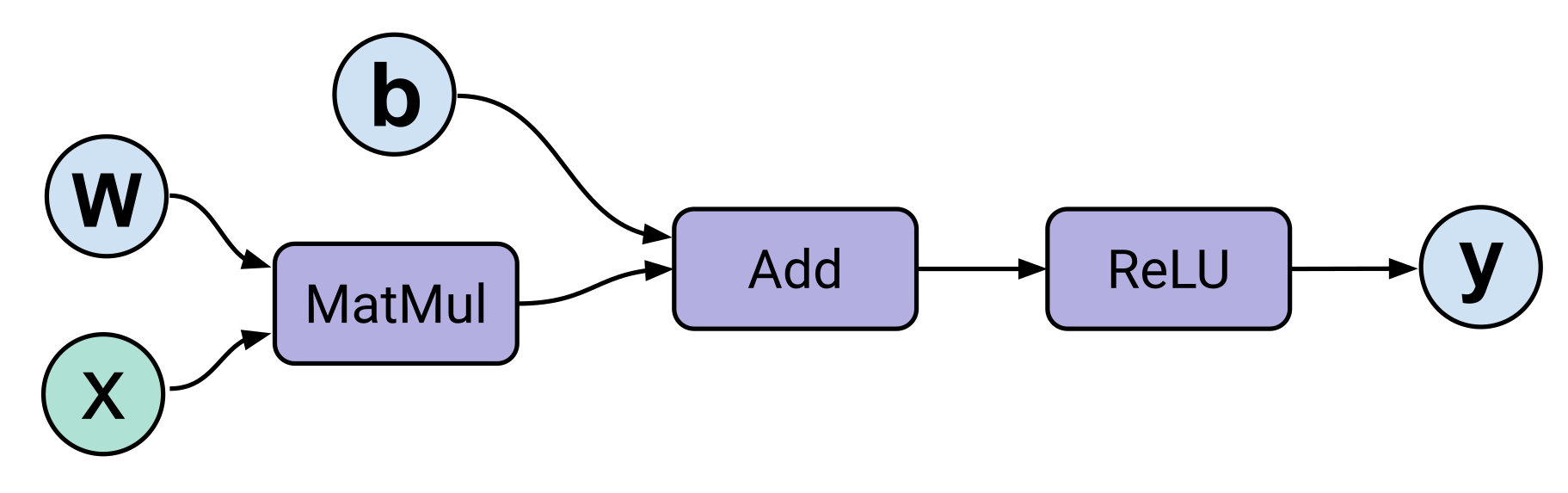}
  \caption[Single perceptron as a computation graph]{The operations shown in a rounded-rectangular box are the nodes of the computation graph which take an arbitrary number of inputs, performs a computation at the node and produces an output. The circular nodes represent the input nodes for tensors. The directed edges show the flow of tensors through the graph.}
  \label{fig:bg:perceptron}
\end{figure}

Computation graphs are a way to graphically represent both the individual tensor operations in a model, and the connections (or data-flow) along the edges between nodes in the graph. Figure \ref{fig:bg:perceptron} shows how the expression, $y = \texttt{ReLU}(\mathbf{w} \cdot \mathbf{x} + b)$, can be represented graphically as a computation graph.

Similarly, a neural network model can be converted into a stateful dataflow (computation) graph in this manner. Using a computation graph as an intermediate representation provides two benefits compared to using a raw model definition. First, we can execute the model on any hardware device as the models have a single, uniform representation that can be modified as required. Second, it allows for pre-execution optimisations based on the host device. For example, we may perform different optimisations for executing on a GPU compared to a TPU (Tensor Processing Unit) requires different data layouts and optimisations.

\subsection{Reinforcement Learning}
Reinforcement learning (RL) is a sub-field of machine learning. Broadly, it aims to compute a control policy such that an agent can maximise its cumulative reward from the environment. Reinforcement Learning has powerful applications in systems environments where a unified model describing the semantics of a system is not available. The agent must itself discover the optimal strategy by a single reward signal.

Formally, RL is a class of learning problem that can be framed as a Markov decision process (MDP) when the MDP that describes the system is not known \cite{bellman1957}; they are represented as a 5-tuple $\langle \mathcal{S}, \mathcal{A}, \mathcal{P}_a, \mathcal{R}_a, \rho_0 \rangle$ where:

\begin{itemize}
  \item $\mathcal{S}$, is a finite set of valid states
  \item $\mathcal{A}$, is a finite set of valid actions
  \item $\mathcal{P}_a$, is the transition probability function that an action $a$ in state $s_t$ leads to a state $s'_{t+1}$
  \item $\mathcal{R}_a$, is the reward function, it returns the reward from the environment after taking an action $a$ between state $s_t$ and $s'_{t+1}$
  \item $\rho_0$, is the starting state distribution
\end{itemize}

We aim to compute a policy, denoted by $\pi$, that when given a state $s \in \mathcal{S}$, returns an action $a \in \mathcal{A}$ with the optimisation objective being to find a control policy $\pi^*$ that maximises the \textit{expected reward} from the environment as can be seen in (\ref{expectedreward}). Importantly, we can control the `far-sightedness' of the policy by tuning the discount factor $\gamma \in [0, 1)$. As $\gamma$ tends to 1, the policy will consider the rewards further in the future but with a lower weight as the distant expected reward may be an imperfect prediction \cite{bellman1957}.

\begin{equation}
  \label{expectedreward}
  \pi^* = \argmax_\pi~\mathbb{E} \left[ \sum^\infty_{t=0} \gamma^t~\mathcal{R}_t \right]
\end{equation}

Classic RL problems are formulated as MDPs in which we have a finite state space. However, such methods quickly become inefficient with large state spaces for applications such as Atari \cite{mnih2013playing, kaiser2020modelbased} and Go \cite{silver2017mastering}. Therefore, we take advantage of modern deep learning function approximators, such as neural networks, which makes learning the solutions far more efficient in practice. We have seen many successful applications in a wide range of fields, for example, robotic control tasks \cite{openai2019solving}, data centre power management, device placement \cite{addanki2019placeto, mirhoseini2018hierarchical}, and playing both perfect and imperfect information games to a super-human level \cite{silver2016mastering,silver2017mastering}. Reinforcement learning excels when applied to environments in which actions may have long-term delayed rewards and inter-connected dependencies that are difficult to learn or model with traditional machine learning techniques.

\subsection{Motivation for Model-based RL}
Model-free and model-based are the two main approaches to reinforcement learning; with recent work such as \cite{app10196685, kaiser2020modelbased, robine2021smaller}, the distinction between the two is becoming somewhat nebulous. It is possible to use a hybrid approach that aims to improve the sample efficiency of the agent by training model-free agents directly in the imagined environment.

The primary benefit of model-based RL is that it has greater sample efficiency, meaning, the agent requires in total fewer interactions with the real environment than the model-free counterparts. If we can either provide or learn a model of the environment that enables the agent to plan actions an arbitrary number of steps ahead, the agent selects from a range of trajectories by taking specific actions to maximise its reward. The agent that acts in this ``\textit{imagined}'' or ``\textit{hallucinogenic}'' environment can be a simple MLP \cite{ha2018worldmodels} to a model-free agent trained using modern algorithms such as PPO \cite{schulman2017proximal}, A2C \cite{mnih2016asynchronous} or Q-learning \cite{watkins1992q, mnih2013playing}. Further, training an agent in the world model is comparatively cheap, especially in the case of complex systems environments where a single episode can be on the order of hundreds of milliseconds.

Unfortunately, learning a model of the environment is not trivial. The most challenging problem that must be overcome is that if the model is imperfect, the agent may learn to exploit the model's deficiencies, thus the agent fails to achieve high performance in the real environment. Consequently, learning an invalid world model can lead to the agent performing actions that may be invalid in an environment where not all actions are valid in all states.

Model-based RL approaches have been applied in a range of environments such as board games, video games, systems optimisation and robotics with a high degree of success. Despite the apparent advantages of model-based RL with regards to reduced computation time, model-free reinforcement learning is by far the most popular approach and massive amounts of compute. Typically, the models are trained on distributed clusters of GPUs/TPUs; large amounts of computing resources are required to overcome the sample inefficiency of model-free algorithms.

\section{RLFlow}

In this section, we introduce RLFlow, a deep learning graph optimiser that uses Reinforcement Learning for automation of graph substitutions, specifically using model-based reinforcement learning by training an agent inside a World Model \cite{ha2018worldmodels} that itself is trained to model the way in which a computation graph is transformed using sub-graph transformations.

\subsection{Reinforcement Learning Formulation}

\subsubsection{System environment}
\label{sec:prob:subsec:sysenv}

In order to train a reinforcement learning agent, it is necessary that we have access to an environment that, given the current environment state, the agent can take an action. After taking the chosen action, the environment is updated into a new state and the agent receives a reward signal. Typically, one uses a mature environment such as OpenAI Gym \cite{brockman2016openai} or OpenSpiel \cite{LanctotEtAl2019OpenSpiel} as the quality of the environment often has a significant impact on the stability of training. Moreover, using an environment that uses a common interface allows researchers to implement algorithms with ease, and importantly, reproduce results from prior work.

In this work, we designed an environment that follows the OpenAI Gym API standard stepping an environment, that is, we have a function \texttt{step(action)} that accepts a single parameter, the action requested by the agent to be performed in the environment. The \texttt{step} function returns a 4-tuple \texttt{(next\textunderscore state, reward, terminal, extra\textunderscore info)}. \texttt{extra\textunderscore info} is a dictionary that can store arbitrary data.

\begin{figure}[ht]
  \centering
  \includegraphics[width=\columnwidth]{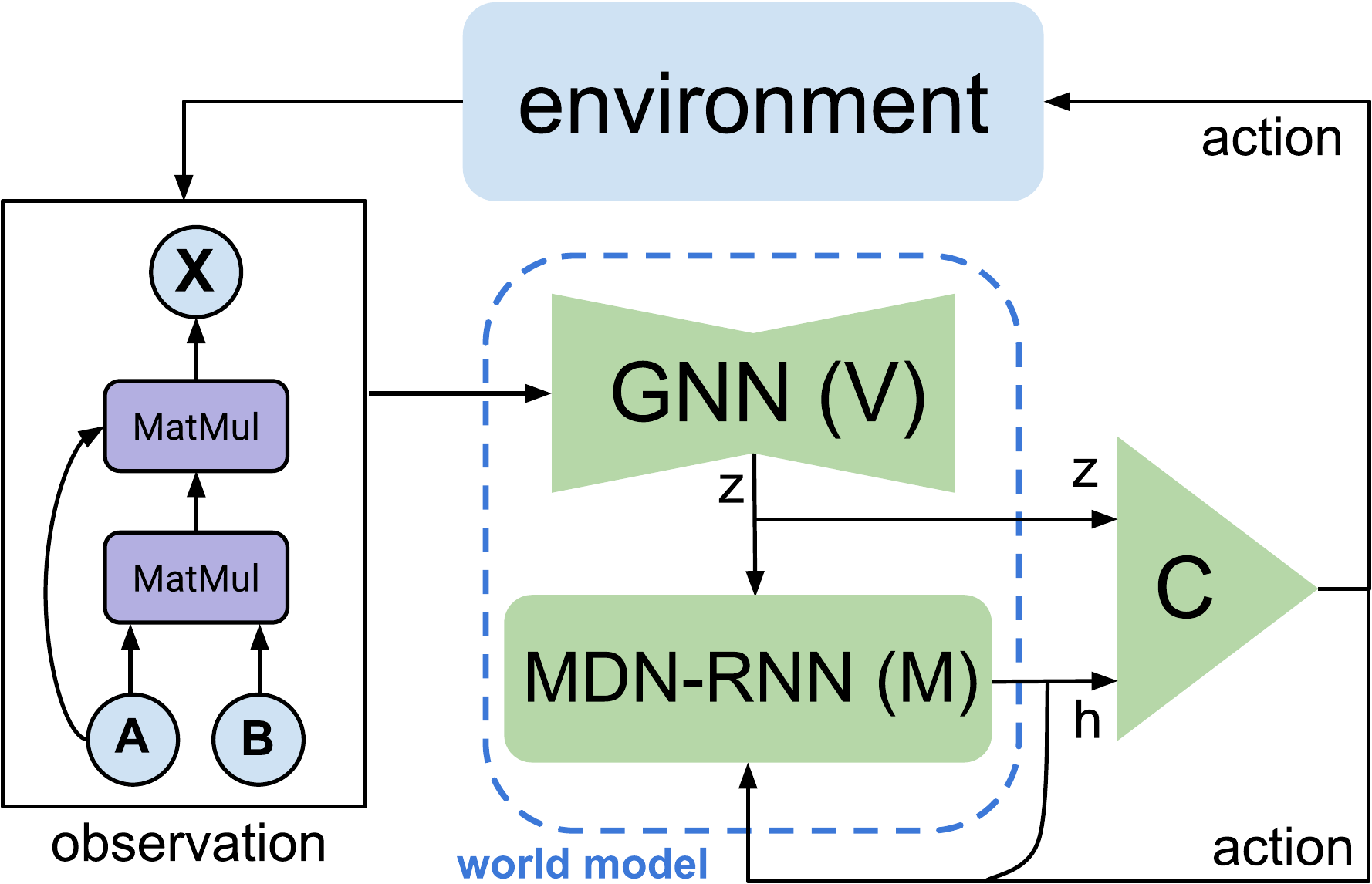}
  \caption{Data flow between components of the system. We select an action given the current state of the imagined environment in the World Model.}
  \label{fig:problem:sys-env}
\end{figure}

We made use of the work by Jia et al. \cite{jia2019taso} who provided an open-source version of TASO as part of the backend of the environment. We provide a computation graph, as well as the chosen transformation and location tuple to TASO. TASO then applies the requested transformation and returns the newly transformed graph. Further, we use internal TASO functions that calculate estimates of the runtime on the hardware device which we use as our reward signal for training the agent. During our experiments we modified TASO to extract detailed runtime measurements to analyse the rewards using a range of different reward functions---we provide more detail in Section \ref{sec:prob:subsec:rwd}.

\subsubsection{Computation Graphs}
The first step prior to optimising a deep learning graph is that we must load, or create on-demand, the model in a supported deep learning framework. In our project, we can support any model that can be serialised into the ONNX \cite{bai2019onnx} binary format which is an open-source standard for defining the structure of deep learning models. By extension, we can support any deep learning framework that supports serialisation of models into the ONNX format such as TensorFlow \cite{tensorflow2015whitepaper, 199317}, PyTorch \cite{paszke2019pytorch} and MXNet \cite{chen2015mxnet}.

Next, we parse the ONNX graph representation by converting all operators into the equivalent TASO tensor representations such that we can modify the graph using the environment API as we described in Section \ref{sec:prob:subsec:sysenv}. Although our environment does not support the conversion of all operators defined in the ONNX specification \footnote{ONNX operator specification:~\url{https://github.com/onnx/onnx/blob/master/docs/Operators.md}}, the majority of the most common operators are supported. Therefore, we still maintain the semantic meaning and structure of the graph. Additionally, after performing optimisations of the graph, we can export the optimised graph directly to an ONNX format.

\subsubsection{State-Action space}
\label{sec:prob:subsec:sap}
In this project we modelled the state and action space in accordance with prior research, specifically we referenced work in the similar domain of system optimisation using reinforcement learning. Mirhoseini et al. \cite{mirhoseini2018hierarchical} used hierarchical RL with multiple actions to find the optimal device placement and Addanki et al. \cite{addanki2019placeto} that also aided in the choice of input/output graph sizes.

Next, we require two values in order to update the environment. First, we need a select a transformation (which we refer to as an \texttt{xfer}) to apply to the graph. Secondly, the location at which to apply the transformation. As we need to select two actions that are dependent on each other to achieve higher performance, it requires selecting the actions simultaneously.

However, doing so would require a model output of $N \times L$ values, where $N$ is the number of transformations, $L$ is the number of locations. Such an action space is too large to train a model to efficiently predict the correct action. Additionally, after choosing a transformation, we ideally mask the available locations as not all locations can be used to apply any transformation. Therefore, using the same trunk network, we first predict the transformation, apply the location mask for the selected transformation, then predict the location.

We define the action as a 2-value tuple of (xfer\textunderscore id, location). There is a special case for the xfer\textunderscore id. When it equals N (the number of available transformations), we consider it the NO-OP action. Therefore, in this special case, we do not modify the graph, rather we terminate the current episode and reset the environment to its initial state.

As explained in the previous section, we used a step-wise approach where at each iteration, we provide a 2-tuple of the transformation and location, to apply in the current state. The updated state from the environment is a 4-tuple consisting of \texttt{(graph\textunderscore tuple, xfer\textunderscore tuples, location\textunderscore masks, xfer\textunderscore mask)}.

\texttt{xfer\textunderscore mask} refers to a binary mask that indicates the valid and invalid transformations that can be applied to the current computation graph as not every transformation can be applied to every graph. If the current graph has only four possible transformations that can be applied, all other transformations are considered to be invalid. Thus, we return a boolean location mask where only valid transformations are set to 1, or \texttt{true}. This can be used to zero out the model logits of invalid transformations (and thereby actions also) to ensure the agent always selects a valid transformation from the set.

Similarly, for each transformation selected by the agent, there are a number of valid locations where this transformation can be applied. We set a hardcoded, albeit configurable, limit the number of locations to 200 in this work. If the current graph has fewer than 200 possible locations for any given transformation, the remaining are considered invalid. Therefore, we again return a boolean location mask, which is named \texttt{location\textunderscore masks} in the 4-tuple defined above, which can be used to zero out the model logits where the locations are invalid. 


\subsubsection{Reward function}
\label{sec:prob:subsec:rwd}
The design of a reinforcement learning agent consists of three key elements, the agent, environment and reward function. Most importantly, we require a reward function that captures dynamics of the environment in such a way that we can directly indicate to the agent if we consider the action to be ``good'' or ``bad''. For example, we wish to prevent the agent from performing actions that would be invalid in the environment, therefore, using the reward signal we provide a large negative reward to disincentivise the agent from replicating the behaviour. Conversely, we need to provide a positive reward, that is calculated using the chosen action and its impact on the agent performance.

Selecting optimal actions can be challenging in any deep reinforcement learning system, especially those with either long-term action dependencies or a large number of possible actions in any given state. Importantly, in our environment, the selection of a poor action be impactful on both subsequent action space and the resulting reward generated by the environment. Therefore, we used multiple reward functions to investigate the resulting performance of the agent. First, we used a simple reward function that is commonly used in sequential RL applications:

\begin{equation}
\label{eq:rwd1}
    r_t =
    \begin{cases}
      RT_{t-1} - RT_t, & \text{if valid action}\\
      \text{-}100,            & \text{otherwise}
    \end{cases}
\end{equation}

Using the reward function defined above, we use the estimated runtime from the prior timestep, $RT_{t-1}$ of the computation graph and the estimated runtime of the current graph, $RT_t$, to determine the step-wise, incremental change in graph runtime as the reward. This simple, yet powerful function has the benefit of very low overhead as we only need to store the last runtime. Furthermore, as our primary goal is to reduce the execution time of the graphs, rather than, for example, the system memory, it directly captures our desired metric which we wish to optimise.

Secondly, we instrumented TASO to extract detailed metrics to design a more complex reward function. We used the runtime, FLOPS, memory accesses, and kernel launches to perform experiments to determine if using a combination of the metrics could yield a higher performance RL agent. We defined the modified reward function as shown below. Where $RT_t$ is the graph runtime at timestep $t$, $M_t$ is the memory accesses at timestep $t$, $\alpha$ and $\beta$ are two hyperparameters for weighting the runtime and memory accesses, respectively.

\begin{equation}
\label{eq:rwd2}
    r_t =
    \begin{cases}
      \alpha(RT_{t-1} - RT_t) + \beta(M_{t-1} - M_t), & \text{if valid action}\\
      \text{-}100,            & \text{otherwise}
    \end{cases}
\end{equation}

We provide further discussion and motivation for our chosen reward functions in Section \ref{sec:eval:subsec:mf:subsubsec:rwd-func} as well as an analysis of the detailed runtime metrics and the impact on graph runtime.

Finally, we note that TASO used a simple method to estimate the runtime of tensor operators that are executed using low-level CUDA APIs and the runtime is averaged over $N$ forward passes. However, this approach to runtime estimation is imperfect as there is a non-negligible variance of the runtime on real hardware and can lead to a poor estimation of the hardware impact. As such, we investigated the use of real runtime measurements during training rather than an estimation of operator runtime. After performing experiments with a modified version of TASO which averages the real runtime over $N$ rounds, we found that it increases the duration of each training step to such a degree that any possible performance improvements achieved using real hardware costs are not worth the trade-off.

\subsection{Graph-level optimisation}

Performing optimisations at a higher, graph-level means that the resulting graph is---in terms of execution methodology---no different than the original graph before optimisation. By performing graph-level optimisation, we generate a graph representation independent of both the platform and backend. The intermediate representation can be optimised by specialised software for custom hardware accelerators such as GPUs and TPUs.

Next, we define that two computation graphs, $\mathcal{G}$ and $\mathcal{G}'$ are semantically equivalent when $\forall \mathcal{I} : \mathcal{G}(\mathcal{I}) = \mathcal{G}'(\mathcal{I})$ where $\mathcal{I}$ is an arbitrary input tensor. We aim to find the optimal graph $\mathcal{G}^*$ that minimises a given cost function, \texttt{cost}$(\mathcal{G})$, by performing a series of transformations to the computation graph at each step, the specific transformation applied does not need to be strictly optimal. In fact, by applying optimisations that reduce graph runtime we further increase the state space for the search; a large state space is preferable in the reinforcement learning domain.

A problem in graph-level optimisation is to define a set of varied, applicable transformations that can be used to optimise the graphs. As previously noted, prior work such as TensorFlow use a manually defined set of transformations and optimise greedily. On the other hand, TASO uses a fully automatic method to generate candidate transformations by performing a hash-based enumeration over all possible DNN operators that result in a semantically equivalent computation graph.

\begin{figure}[htp]
  \centering
  \subcaptionbox{Tensor renaming substitution \label{fig:problem:rewrite-graph1}}{\includegraphics[width=0.40\columnwidth]{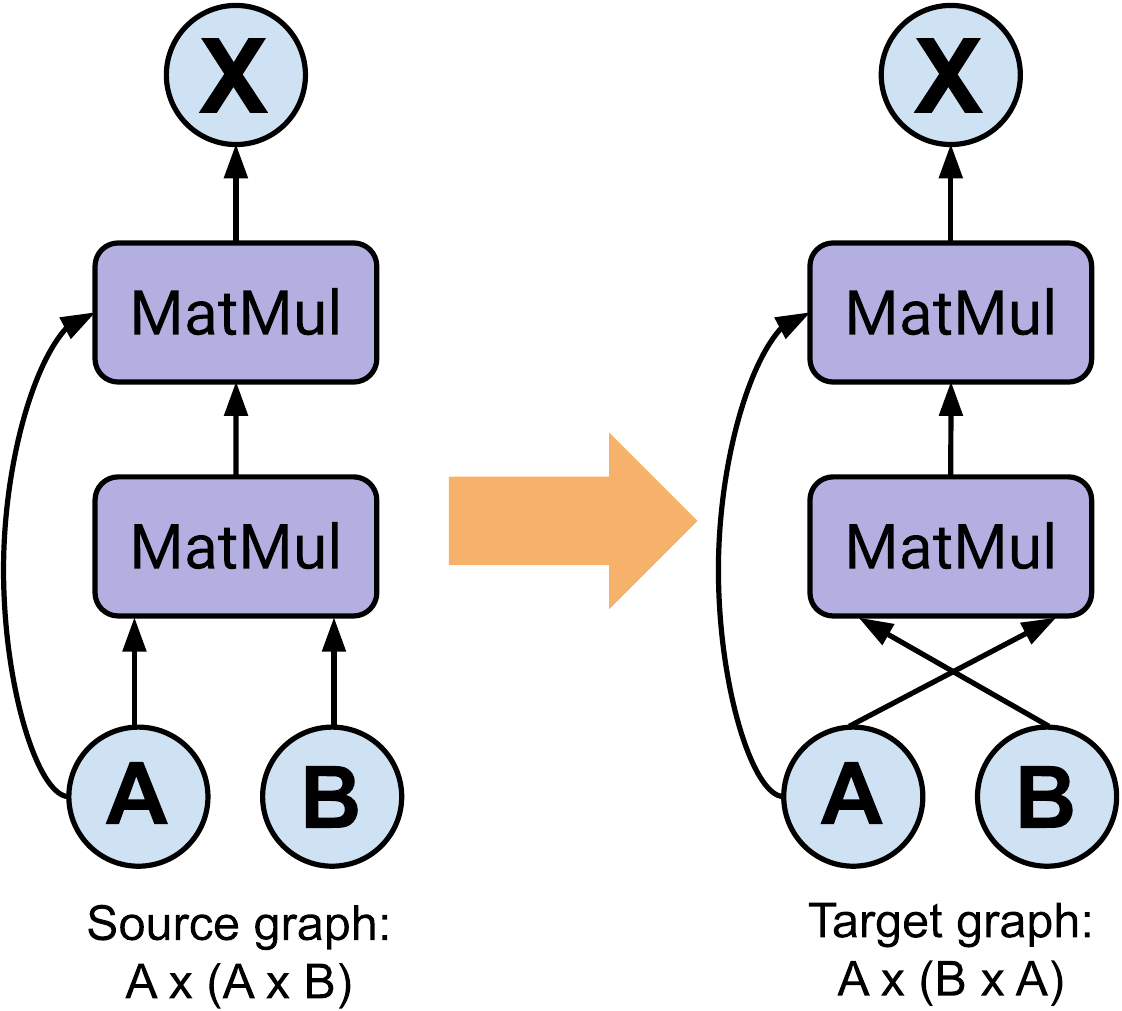}}%
  \hfill
  \subcaptionbox{Common subgraph substitution \label{fig:problem:rewrite-graph2}}{\includegraphics[width=0.54\columnwidth]{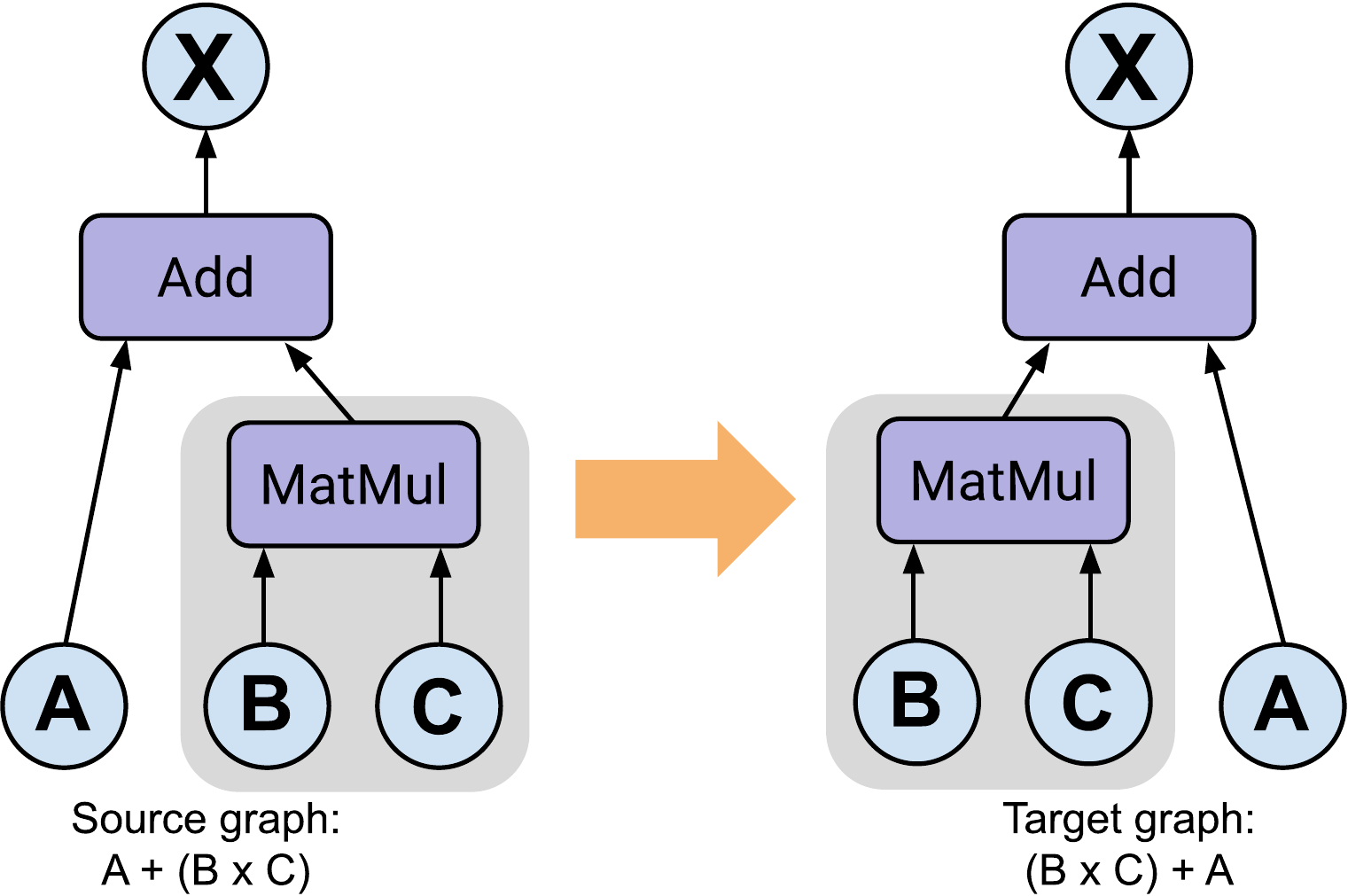}}%
  \caption{Two examples of trivial graph substitutions that does not impact the overall runtime of the computation graph. The left sub-figure shows a simple renaming of the tensor inputs. The figure on the right shows that we have a common sub-graph between the source and the target graphs. In both cases we eliminate the duplicates as the hash of the two graphs will be identical.}
\end{figure}

In this work, we take the same approach as TASO and automatically generate the candidate sub-graphs. We perform this as an offline step as it requires a large amount of computation to both generate and verify the candidate substitution; to place an upper bound on the computation, we limit the input tensor size to a maximum of 4x4x4x4 during the verification process. Following the generation and verification steps, we prune the collection to remove substitutions that are considered trivial and, as such, would not impact runtime. For example, trivial substitutions include input tensor renaming and those with common sub-graphs. We show both techniques diagrammatically in Figure \ref{fig:problem:rewrite-graph1} and \ref{fig:problem:rewrite-graph2} respectively.

\subsection{World Models}
World models, introduced by Ha et al. \cite{ha2018worldmodels}, create an imagined model of the true environment by observing states, actions and rewards from the environment and learning to estimate the transitions between states based upon the actions taken. Ha et al. showed that the world models can learn the environment transitions and achieve high performance on visual learning tasks such as CarRacing and VizDoom that exceeds model-free agents. One should note that Ha \& Schmidhuber used RGB pixel images as input to a convolutional neural network to generate the latent space embedding. In comparison, we used the latent space produced by the graph neural network using the graph as input. In either case, we aim to learn the world model using the latent space from the environment.

\subsubsection{Recurrent Neural Networks}
\label{sec:rlopt:subsec:rnn}

Recurrent Neural networks (RNNs) are a class of architectures in which the connections between the nodes form a directed graph in a temporal sequence \cite{650093}. Importantly, as the output of an RNN is deterministic, we use the outputs from the RNN as the parameters for a probabilistic model to insert a controllable level of stochasticity in the output predictions; a method first proposed by Graves \cite{graves2014generating}.

\begin{figure}[htp]
  \centering
  \includegraphics[width=\columnwidth]{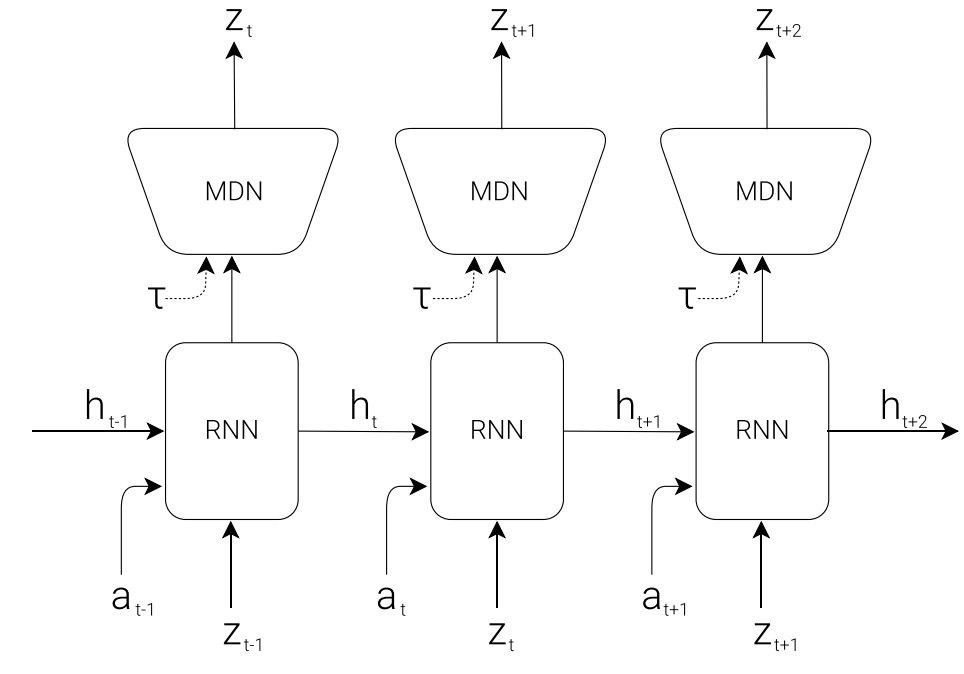}
  \caption[Temporally unrolled MDN-RNN]{Structure of an unrolled MDN-RNN. The MDN-RNN generates an output which is used as the parameters of a Gaussian mixture distribution. The parameterised GMM is then used to sample the distribution for the vector $z_{t+1}$; the MDN is controlled by the temperature parameter $\tau$. Adapted from \cite{ha2018worldmodels}.}
  \label{fig:rl:mdnrnn}
\end{figure}

A constraint of using RNNs is that they expect a fixed-sized input sequence. In this work, the shape of the latent state tensor and the number of actions available to the agent in a rollout is variable. We use a common approach, called padding, to mitigate this problem by padding the input sequence with zero values until the desired length. After performing inference on the model and retrieving the predicted state, we mask the results based on the input padding to ensure we only use valid predictions to select the next action using the controller.

\subsubsection{MDN-RNN}

By combining the mixture density and recurrent networks, we can use rollouts of the environment sampled using a random agent to train the combined network, called an MDN-RNN. We use the network to model the distribution $P(z_{t+1}~|~a_t, z_t, h_t)$, where $z_t$ and $z_{t+1}$ is the latent state at the times $t$ and $t+1$ respectively, $a_t$ is the action performed at time $t$, and $h_t$ is the RNN networks hidden state at time $t$. Figure \ref{fig:rl:mdnrnn} shows the combination of the RNN and MDN networks and how we calculate the predictions of the next latent state in sequence.

Furthermore, after training the world model, we must train an agent (or controller) to perform actions in the world model that learns to take optimal actions to maximise reward. For inference of the world model, a softmax layer is used that outputs $\pi$ in the form of a categorical probability distribution that we sample under the Gaussian model, often referred to as a Gaussian Mixture Model (GMM), parameterised by $(\mu_i, \sigma_i)$.

In Figure \ref{fig:rl:mdnrnn} we show that one of the inputs to the MDN is $\tau$, the temperature. By altering the temperature, it allows us to control the stochasticity of the agent during the training of the controller. The logits of the RNN that represent the predictions for the values of $\pi$ are divided by the temperature prior to being passed into the softmax function which converts the logits into pseudo-probabilities. We incorporate the temperature, $\tau$, into the function using the following equation.

$$
\text{softmax}(\mathbf{x}_i) = \frac{\exp\left( \frac{x_i}{\tau} \right) }{\Sigma_j \exp \left( \frac{x_j}{\tau} \right) }
$$

Typically, the temperature is a real number in the range $\tau \in \left[ 0, 1 \right)$, where a value of zero leads to completely deterministic predictions generated by the RNN, whereas larger values introduce a greater amount of stochasticity in predicted values. An increasing value of $\tau$ increases the probability of selecting samples with a lower likelihood; leading to a greater diversity of actions taken by the agent in the environment. Importantly, Ha et al. \cite{ha2018worldmodels} found that having a large temperature can aid in preventing the agent from discovering strategies to exploit in the world model that is not possible true system environment due to an imperfect model.

Modifying the softmax activation function in this way is equivalent to performing knowledge distillation between two models; learnt information is transferred from a large teacher model, or ensemble model, to a smaller model which acts as a student model \cite{hinton2015distilling}. In both the context of knowledge distillation and training a controller network using the world model, a high temperature will generate softer targets. Specifically, in this work, a higher temperature produces a softer pseudo-probability distribution for $\pi$ in the GMM. Additionally, using soft targets will provide a greater amount of information for the model to be learnt by forcing the model to learn more aggressive policies, thus outputting stochastic predictions that is beneficial to encourage exploring the environment state-action space.

Furthermore, we consider how the world model is trained. For any supervised learning task, we require target data to compare our predictions against, calculate a loss and perform backpropagation to update the weights in the network. To train the world model, we use a random agent. The probability of the agent choosing any action from the set of valid actions is equal. Unlike Ha et al. \cite{ha2018worldmodels} who performed 10,000 rollouts of the environment offline using a random policy to collect the data, we took a different approach.

Rather than generating large rollouts offline, we generated minibatch rollouts using the random agent online and directly used the observations to train the world model. Although this approach reduces the data efficiency as we only use each state observation once, we benefit from removing the need to generate the data prior to training. In systems environments, it is often expensive---in terms of computation time---to step the environment to collect a diverse dataset. Therefore, we found generating short rollouts and training on the minibatch was beneficial without any perceivable impact on performance.

\subsection{Action Controller}
\label{sec:rlopt:subsec:actionctrl}

Finally, we discuss the design of the ``controller'', the network/agent that learns to output actions based upon the output from the MDN-RNN world model. Ha and Schmidhuber \cite{ha2018worldmodels} used an evolution-based controller defined as a simple multi-layer-perceptron, $a_t = W_c[z_t, h_t] + b_c$, that accepts the hidden and current states from the recurrent network to predict the next action to be taken. When training the controller inside the world model environment, we no longer have access to the ground truth state nor the reward produced by the real environment. Therefore, we cannot use supervised learning to train the controller. 

In \cite{ha2018worldmodels} the authors used an evolutionary algorithm, co-variance matrix adoption evolution strategy (CMA-ES) \cite{hansen2001completely, hansen2016cma}, which optimises the weights of the network based on the reward produced by the world model. Alternatively, recent work by Hafner et al. \cite{hafner2020dream, hafner2021mastering} has shown to achieve state-of-art results in the Atari environment using an actor-critic method as the controller in the world model. Furthermore, prior work on the application of world models to systems environments has shown one can train a model-free controller inside the world environment \cite{app10196685}.


\section{Evaluation}

In this section, we evaluate the aims described at the beginning of the paper. We claimed to use reinforcement learning to perform automated optimisation of deep learning computation graphs. Thus, this evaluation seeks to answer the following questions:

\begin{enumerate}

\item Are model-based reinforcement learning methods able to model the transition dynamics of the environment?

\item Are the agent policies able to generalise to unseen states of the same graph to act in accordance to our performance objectives?

\item Do the world models accurately model the reward estimation from the graphs latent state?

\item Are the agents trained in an imagined world model applicable to the real-world environment?

\item Is RL approach more effective for the sequence to sequence models (e.g. Transformers)?

\end{enumerate}

Finally, we conclude with an overall discussion of our findings and their impact.

\subsection{Experimental Setup}

All experiments presented were performed using a single machine running Ubuntu Linux 18.04 with a 6-core Intel i7-10750H@2.6GHz, 16GB RAM and an NVIDIA GeForce RTX 2070.

To interface with the internal representation of the computation graphs, as previously discussed, we used the open-sourced version of TASO \cite{jia2019taso} which we modified to extract detailed runtime information. Further, we implemented the reinforcement learning algorithms in TensorFlow \cite{tensorflow2015whitepaper, 199317} and utilised the \texttt{graph\textunderscore nets} package developed by Battaglia et al. \cite{battaglia2018relational} to process our input graphs using the method which we described in Section \ref{sec:prob:subsec:sysenv}. The PPO agent was implemented based upon the implementation provided by Schulman et al. \cite{schulman2017proximal}.

\subsection{Graphs Used}
\label{sec:eval:subsec:graphsused}

\begin{table}[htb!]
  \centering
  \resizebox{\columnwidth}{!}{
    \begin{tabular}{@{}ccccc@{}}
      \toprule
      Graph         & Type          & Layers & Unique Layers & Substitutions \\ \midrule
      InceptionV3   & Convolutional & 43     & 12            & 56            \\
      ResNet-18     & Convolutional & 18     & 6             & 40            \\
      ResNet-50     & Convolutional & 50     & 6             & 228           \\
      SqueezeNet1.1 & Convolutional & 21     & 3             & 288           \\
      BERT-Base     & Transformer   & 12     & 3             & 80            \\
      ViT-Base      & Transformer   & 16     & 5             & 60            \\ \bottomrule
      \end{tabular}
  }
  \caption[Properties of evaluation graphs]{Properties of the six evaluation graphs used in the experiments contained in this chapter. We differentiate the total number of layers in a network from the number of unique layers used in composing the network to provide a more accurate representation of its complexity.}
  \label{table:eval:graph-props}
\end{table}

We chose to use six real-world deep learning models to evaluate our project. InceptionV3 \cite{szegedy2015rethinking} is a common, high-accuracy model for image classification trained on the ImageNet dataset\footnote{\url{https://image-net.org/index.php}}. ResNet-18 \& ResNet-50 \cite{he2015deep} are also deep convolutional networks that are 18 and 50 layers deep respectively. SqueezeNet \cite{iandola2016squeezenet} is a shallower yet accurate model on the same ImageNet dataset. BERT \cite{devlin2019bert} is a recently introduced large transformer network that has been to improve Google search results \cite{nayak2019}. Finally, ViT, a transformer network specifically designed for computer vision tasks; ViT \cite{dosovitskiy2021image} has been shown to outperform traditional convolutional networks at image recognition tasks. As these graphs were also used in the evaluation of TASO \cite{jia2019taso}, we can show a direct comparison of the performance between the different approaches.

\subsection{Reward functions}
\label{sec:eval:subsec:mf:subsubsec:rwd-func}

As we described in Section \ref{sec:prob:subsec:rwd}, the design of the reward function used in the training of RL agents is a pivotal part of the agent's architecture. In this section, we analyse our proposed reward functions and their effect on the agent's convergence.

Furthermore, we collected detailed runtime measurements that were recorded at each step of the training process to analyse the correlation between metrics. We collected metrics such as runtime, FLOPS, memory access and the number of kernel launches. As one may suspect, we found that the estimated runtime and memory accesses have a strong correlation; when the memory access decreases, we see a notable decrease in estimated runtime. Therefore, by providing reward signals to the agent that combine the effect of runtime or memory access, the agent will learn to optimise for both metrics.


\begin{figure}[htb!]
  \centering
  \includegraphics[width=1\columnwidth]{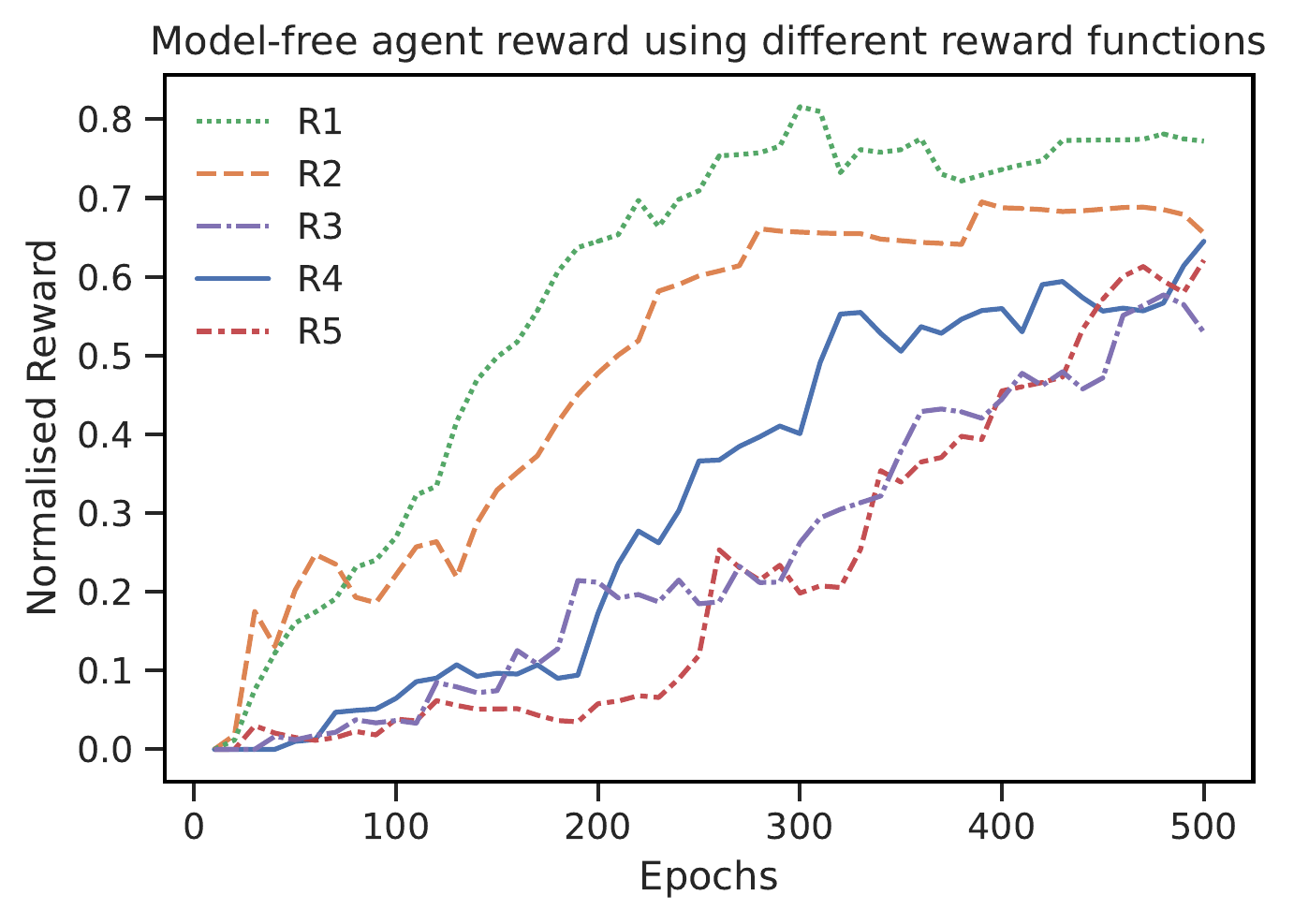}
  \caption[Agent reward using various reward functions]{Normalised reward of each agent using various reward functions while being trained for 500 epochs. R1 uses the second reward function with tuned parameters, R2 uses new runtime reward, R3 uses $\alpha=0.1, \beta=0.9$, R4 uses $\alpha=0.5, \beta=0.5$ and finally, R5 uses incremental runtime improvement.}
  \label{fig:eval:mf-reward-funcs}
\end{figure}

Figure \ref{fig:eval:mf-reward-funcs} shows the effect on the convergence of model-free RL agents while training on the BERT graph using various reward functions. Significantly, using the approach of the first reward function (\autoref{eq:rwd1}) shows that the model-free agent improves at a linear rate as shown by R4 in Figure \ref{fig:eval:mf-reward-funcs}. Alternatively, the second reward function (\autoref{eq:rwd2}) using tuned hyperparameters of $\alpha$ and $\beta$, shown by R1 in Figure \ref{fig:eval:mf-reward-funcs}, converges the fastest.

However, the rate of convergence does not show the whole picture of the agent's performance. Despite having a restricted training schedule of 500 epochs, the highest performing agent was trained using reward function R1. It achieved an average runtime improvement of $48.7 \pm 3.2\%$. Surprisingly, the second-highest performing agent was the agent using R4, the simplest reward function from the set tested, with a performance of $43.2 \pm 2.3\%$.

To find the optimal values of the hyperparameters $\alpha$ and $\beta$ that achieved maximum performance, we performed a grid search for $\alpha$ and $\beta = 1 - \alpha$ between the values of $[0, 1]$ in increments of $0.1$. After training each agent for 500 epochs and evaluating the agent's performance for five runs, we found that the reward function resulting in the highest performance was using the values $0.8$ and $0.2$ for $\alpha$ and $\beta$ respectively.

\subsection{Runtime Performance}
\label{sec:eval:subsec:mb:sec:runtimeperf}

\begin{figure*}[htb!]
  \centering
  \includegraphics[width=\textwidth]{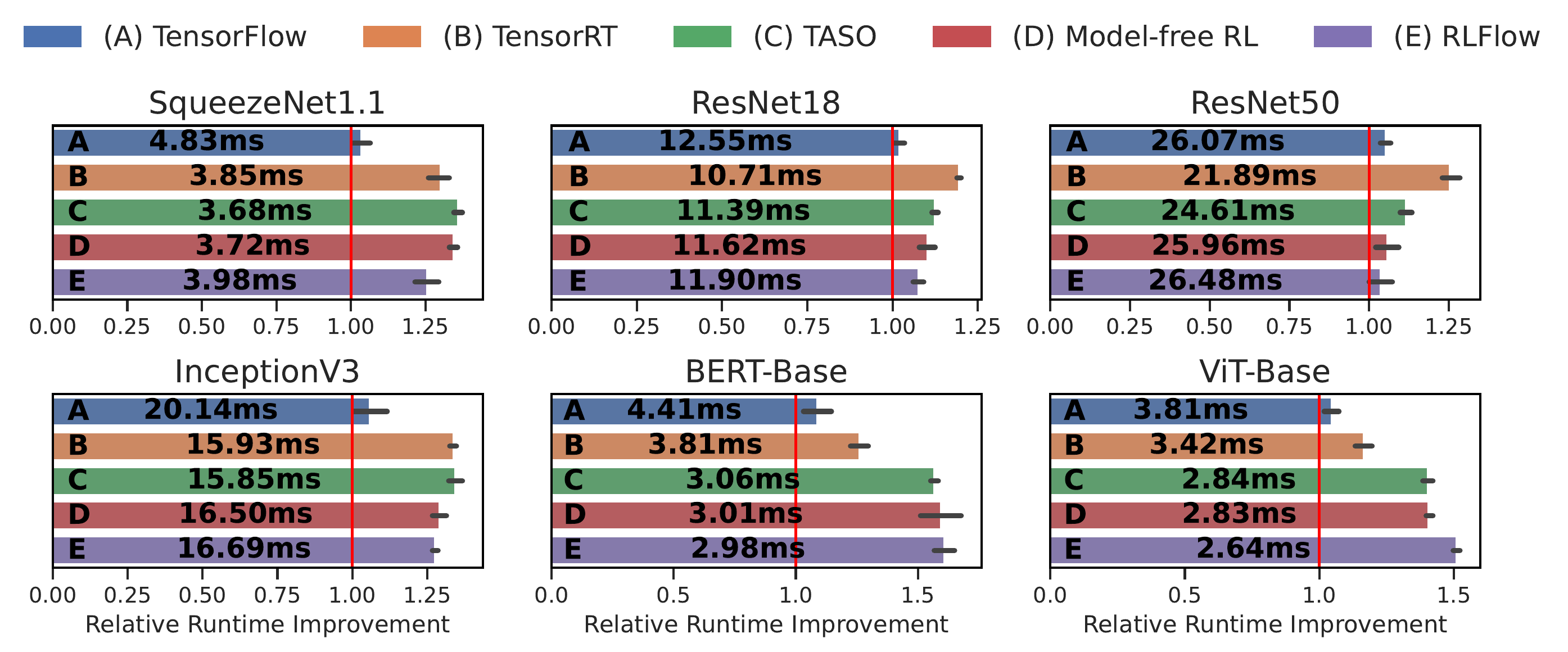}
  \caption{Runtime of optimised graphs using an agent trained using the model-based world model. We also show the baseline results as comparison. The x-axis shows the relative runtime improvement, a higher relative runtime is better. Each experiment was performed five times; we report the mean inference time, and 95\% CI through error bars.}
  \label{fig:eval:world-model-runtimes}
\end{figure*}

Figure \ref{fig:eval:world-model-runtimes} shows the runtime of the optimised graphs for the model-based agents trained inside the fully hallucinogenic world model. Each agent was trained inside a world model using rollouts from the respective graph, as described in Section \ref{sec:rlopt:subsec:actionctrl}. We trained the agents for a maximum of 1000 epochs, in mini-batches of 10 epochs. Furthermore, during training of the controller agent network, we used a fixed learning rate for both the policy and value networks.

Firstly, we note that training the agents on convolutional networks, especially SqueezeNet1.1 and InceptionV3, the model-based agent failed to outperform TASO, although we still decreased the runtime compared to the graph produced by TensorFlow optimisations. Importantly, we observe that the model-based agent outperformed all baseline approaches on the BERT transformer network; we improved the runtime by 54.1\% and 7.3\% compared to TensorFlow and TASO, respectively. Figure \ref{fig:eval:xfer-heatmap} shows the transformations applied by the model-based agent on the test graphs, and compared to TASO, we only apply a single transformation over 20 times---compared to TASO that uses four transformations to produce the optimised graph.

Our model-based agent achieved a similar level of performance on the majority of the tested graphs compared to a model-free agent trained using the real system environment. The model-free agent was trained for 2000 epochs and, by extension, over 4,000,000 interactions with the real environment. Comparatively, the model-based agent performed approximately 1,000,000 interactions with the real environment as the agent did not interact with the real environment while training inside the world model. Therefore, it is evident that by training inside the world model, we improved the sample efficiency of the agent. On the other hand, the agent's performance decreased compared to the model-free agent in four of the six tested graphs.

Furthermore, an important consideration is the wall-clock time for stepping the environment to a new state, based upon the agent action, when training inside a systems environment. We analysed the time required to perform a single step while training on the ResNet50 graph. We found that stepping the world model (performing inference of the world-model) takes, on average 10ms, whereas stepping the real environment takes on average 850ms. Thus, although the performance of the model-based agent was comparatively lower, our wall-clock time required for training was reduced by a factor of 85x.

Finally, we can observe that RLFlow has higher performance when optimising transformer-based models compared to convolutional networks. A transformer is a deep learning model that adopts the mechanism of attention based on Sequence-to-Sequence (Seq2Seq) architecture \cite{cho2014learning, sutskever2014sequence, vaswani2017attention}. Seq2Seq models are particularly good at translation, such as Natural Language Processing (NLP), where the sequence of words from one language is translated into an equivalent sequence of words in another language. A popular choice for this type of model is Long-Short-Term-Memory (LSTM) based models. With sequence-dependent data, the LSTM modules can give meaning to the sequence while remembering, or forgetting, the parts it finds important, or unimportant.

Transformers are getting very popular nowadays. Not only in Natural Language Processing (NLP), Vision Transformers (ViTs) \cite{dosovitskiy2021image} give a good performance on Computer Vision tasks, to the point of outperforming traditional CNNs that have excelled at such tasks. The trend seems to favour Transformers for the majority of NLP tasks. Although, CNNs are still important and are now embedded in GNNs.

\subsection{Optimisation Time}

\begin{figure}[htb!]
  \centering
  \includegraphics[width=1.0\columnwidth]{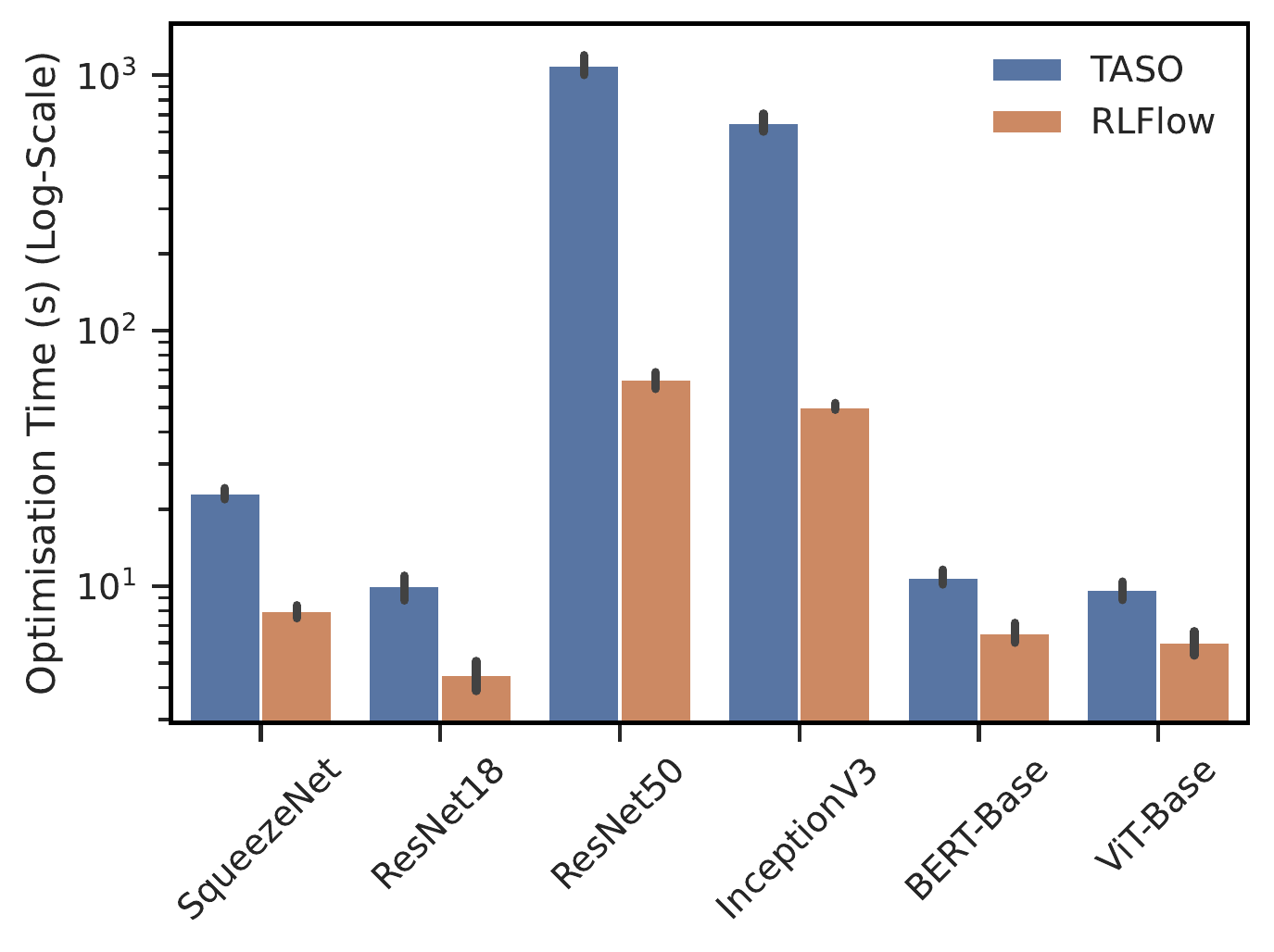}
  \caption[Optimisation time for TASO and MB-RL]{Time required to generate the optimised model using model-based RL and TASO. In all cases, the optimisation time using our proposed approach is lower than that of TASO.}
  \label{fig:eval:optimisation-time}
\end{figure}

Figure \ref{fig:eval:optimisation-time} shows the optimisation time required to generate the optimised graph of the tested models. We note that the optimisation time for the model-based agent does not include the time needed to learn the world model, nor training the controller in the environment. The optimisation time is measured as the wall-clock time to perform $n$ agent steps to generate the optimised graph using our RL agents. Thus, although TASO has a longer optimisation time compared to the RL agent, TASO only requires us to perform the cost-based search a single time.

\subsection{Memory Usage}

\begingroup
\setlength{\tabcolsep}{2pt}
\begin{table}[!t]
  \caption{Relative performance improvement of the graphs when optimised by RLFlow ($\tau = 1.0$) compared to TensorFlow}
  \label{table:eval:graph-mem-usage}
  \centering
    \begin{tabular}{@{}cccccc@{}}
      \toprule
      \multicolumn{1}{l}{} & \multicolumn{2}{c}{TensorFlow}                                         & \multicolumn{2}{c}{RLFlow} \\ \cmidrule(lr){2-3} \cmidrule(lr){4-5}
      \multicolumn{1}{l}{} & \multicolumn{1}{l}{Inf. time (ms)} & \multicolumn{1}{l}{Mem. usage (GiB)} & \multicolumn{2}{c}{\% Improvement}    \\ \midrule
      ResNet18      & 12.55  & 1.18 & 5.2\%  & 1.1\% \\
      ResNet50      & 26.07  & 2.34 & -1.6\% & 0.6\% \\
      InceptionV3   & 20.14  & 2.11 & 17.1\% & 2.3\% \\
      SqueezeNet1.1 & 4.83   & 1.14 & 17.6\% & 1.8\% \\
      BERT-Base     & 4.41   & 0.26 & 32.4\% & 4.5\% \\
      ViT-Base      & 3.81   & 0.34 & 30.7\% & 3.2\% \\ \bottomrule
      \end{tabular}
\end{table}
\endgroup

Table \ref{table:eval:graph-mem-usage} shows the percentage improvement of both the inference time and the memory used for performing inference on the optimised models. Importantly, although we only tasked the agent to optimise for reducing the runtime of the graphs, we observe a secondary effect in a reduction of memory usage by the model of up to 4.5\%, compared to TensorFlow. Although this result is not unsurprising, as optimising a model's architecture leads to a smaller model, based upon the number of trainable parameters. It shows that tasking the agent with reducing runtime does not come at the expense of other resources with strict usage constraints.

\subsection{World-model accuracy}
\label{sec:eval:subsubsec:wm-acc}

\begin{figure}[htb!]
  \centering
  \includegraphics[width=1\columnwidth]{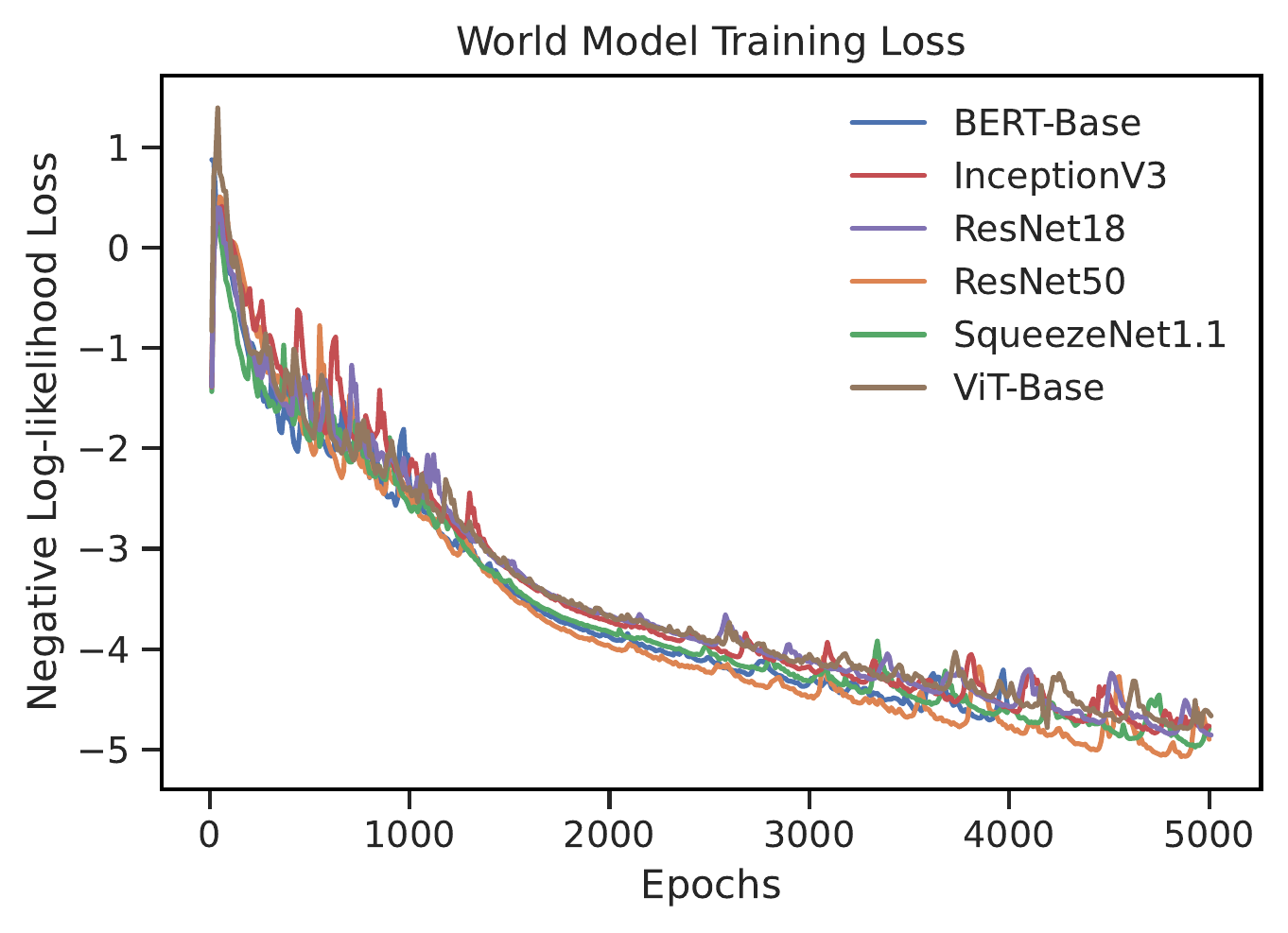}
  \caption[Log-likelihood loss of world models]{Log-likelihood loss during training on each of the six graphs. The learning rate was decayed over the course of 5000 epochs using a 2nd-degree polynomial decay policy.}
  \label{fig:eval:world-model-loss}
\end{figure}

The training of the model-based agent is split into two parts. First, we train the world model, the network that learns to simulate the environment dynamics, and secondly, we train the controller network inside the world model. Figure \ref{fig:eval:world-model-loss} is a plot of the log-likelihood loss for each graph during training of the world model; it shows the convergence of the world model during training after approximately 5000 epochs on all graphs. We used the same hyperparameters for training each world model as well as decaying the learning rate over the course of 5000 epochs with a 2nd-degree polynomial decay policy. We note that despite all the graphs having different architectures, depths and transformation availability, the world model is able to generalise and learn to represent the transformation state transitions accurately after a short period of training. The MDN-RNN is trained with 8 Gaussians and 256 hidden units, all other hyperparameters used in training the MDN-RNN world model are the same as those used by Ha and Schmidhuber \cite{ha2018worldmodels}, unless otherwise stated.

\begin{figure}[htb!]
  \centering
  \includegraphics[width=1\columnwidth]{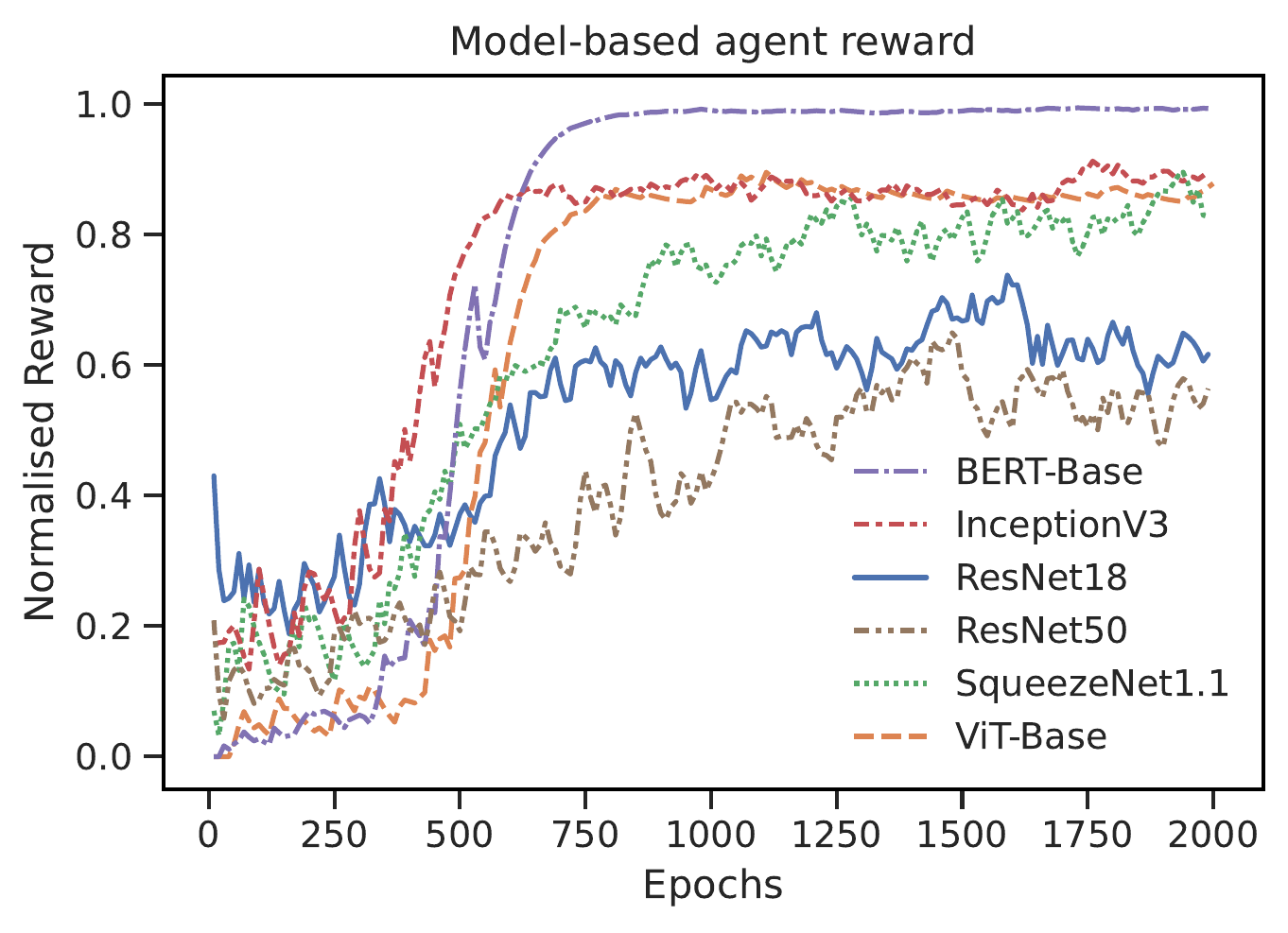}
  \caption[Predicted epoch reward during training of agent in world model]{Predicted reward produced by the world model while training the agent inside a the imagined environment. All rewards are normalised into the same range.}
  \label{fig:eval:world-model-pred-reward}
\end{figure}

Figure \ref{fig:eval:world-model-pred-reward} shows the reward (decrease in estimated runtime) for each graph as predicted by the world model during training. The tested graphs have a wide range of epoch rewards. We perform min-max normalisation to scale the plots into the same range. This finding also supports the results shown in Figure \ref{fig:eval:world-model-runtimes}. The agent applies higher-performing optimisations to transformer networks (BERT-Base and ViT-Base) compared to convolutional networks. Notably, for all transformer networks, the optimal strategy is found after approximately 700 epochs.

On the contrary, networks such as ResNet 18/50 are less stable during training with a high epoch to epoch variation in rewards, and we see that the agent struggles to learn an optimal sequence of subgraph transformations to apply.

If we assume that both the model-free and model-based agents should achieve a similar level of performance once trained, the results in Figure \ref{fig:eval:world-model-pred-reward} shows that the agents trained in the world model are less stable. We hypothesise that there are three factors for such a discrepancy to occur:

\begin{itemize}
  \item Imperfect world-model reward predictions leading to incorrect (or invalid) actions being performed
  \item Next state prediction by the world-model generating states that are invalid due to poor generalisation of the model
  \item Incorrect action mask predictions that would lead to a divergence between the world-model state and real environment state 
\end{itemize}

In an attempt to resolve the issues highlighted above, we performed further experiments that aid in both reducing the variance in reward predictions, as well as stabilise the world model during training to prevent state divergence over time. We performed a temperature sweep of the hyperparameter $\tau$ that is used in training agents inside the world model, shown in Section \ref{sec:eval:subsec:temp-sweep}.

\subsection{Temperature Sweep}
\label{sec:eval:subsec:temp-sweep}

\begin{table}[!htb]
  \caption{Temperature sweep of trained model-based agent optimising the BERT network. We show the percentage improvement of the models compared to TensorFlow in the scores produced by the world model and in the real environment. The highest performing model (over five runs) in shown in \textbf{bold}}
  \label{table:eval:agent-temperatures}

  \centering
  \resizebox{\columnwidth}{!}{
  \begin{tabular}{@{}lll@{}}
  \toprule
  Temperature  & World-model Score        & Real Score                \\ \midrule
  0.1          & 6.67\% $\pm$ 0.6\%          & 43.92\% $\pm$ 5.1\%         \\
  0.5          & 7.75\% $\pm$ 0.3\%          & 55.33\% $\pm$ 6.7\%         \\
  0.75         & 9.10\% $\pm$ 0.4\%          & 55.80\% $\pm$ 5.2\%         \\
  1.0          & 8.85\% $\pm$ 1.2\%          & 55.78 $\pm$ 4.0\%           \\
  1.2          & 9.91\% $\pm$ 0.8\%          & 57.01\% $\pm$ 3.9\%         \\
  \textbf{1.5} & \textbf{10.20\% $\mathbf{\pm}$ 0.6\%} & \textbf{58.23\% $\mathbf{\pm}$ 3.6\%} \\
  1.75         & 9.92\% $\pm$ 1.0\%          & 52.07\% $\pm$ 5.8\%         \\
  2.0          & 9.65\% $\pm$ 0.8\%          & 46.12\% $\pm$ 5.4\%         \\
  2.5          & 10.04\% $\pm$ 2.0\%         & 41.14\% $\pm$ 10.2\%         \\
  3.0          & 10.38\% $\pm$ 1.9\%         & 51.32\% $\pm$ 7.2\%         \\ \bottomrule
  \end{tabular}}
\end{table}

Table \ref{table:eval:agent-temperatures} shows the results from performing a temperature sweep in which we used different values of $\tau$ while training the agent in a world model. After training, we evaluated the agent that produced an optimised graph which we evaluated to determine average runtime. The table shows the average reduction in runtime and standard deviation, averaged over five runs, compared to the unoptimised graph. The motivation for using a range of temperatures is that a higher value of $\tau$ leads to softer targets for the agent to predict, thereby improving generalisation. Conversely, a lower value of $\tau$ presents hard targets and thus, when $\tau = 1.0$, it is equivalent to using the unmodified mixing weight, $\pi$, of the MDN.

Based upon the results in table \ref{table:eval:agent-temperatures} from the conducted experiments, we note that the world model agents are stable to temperatures within the range of $\tau = 0.5$ to $\tau = 1.75$. Although the runtime improvement world-model from the environment is consistently above 6\%, we observe a large difference between the predicted runtime improvement and the real environment reward. One approach to producing more accurate reward predictions could be to use a separate reward prediction network. 

\subsection{Graph transformations}

\begin{figure}[htb!]
  \centering
  \includegraphics[width=1\columnwidth]{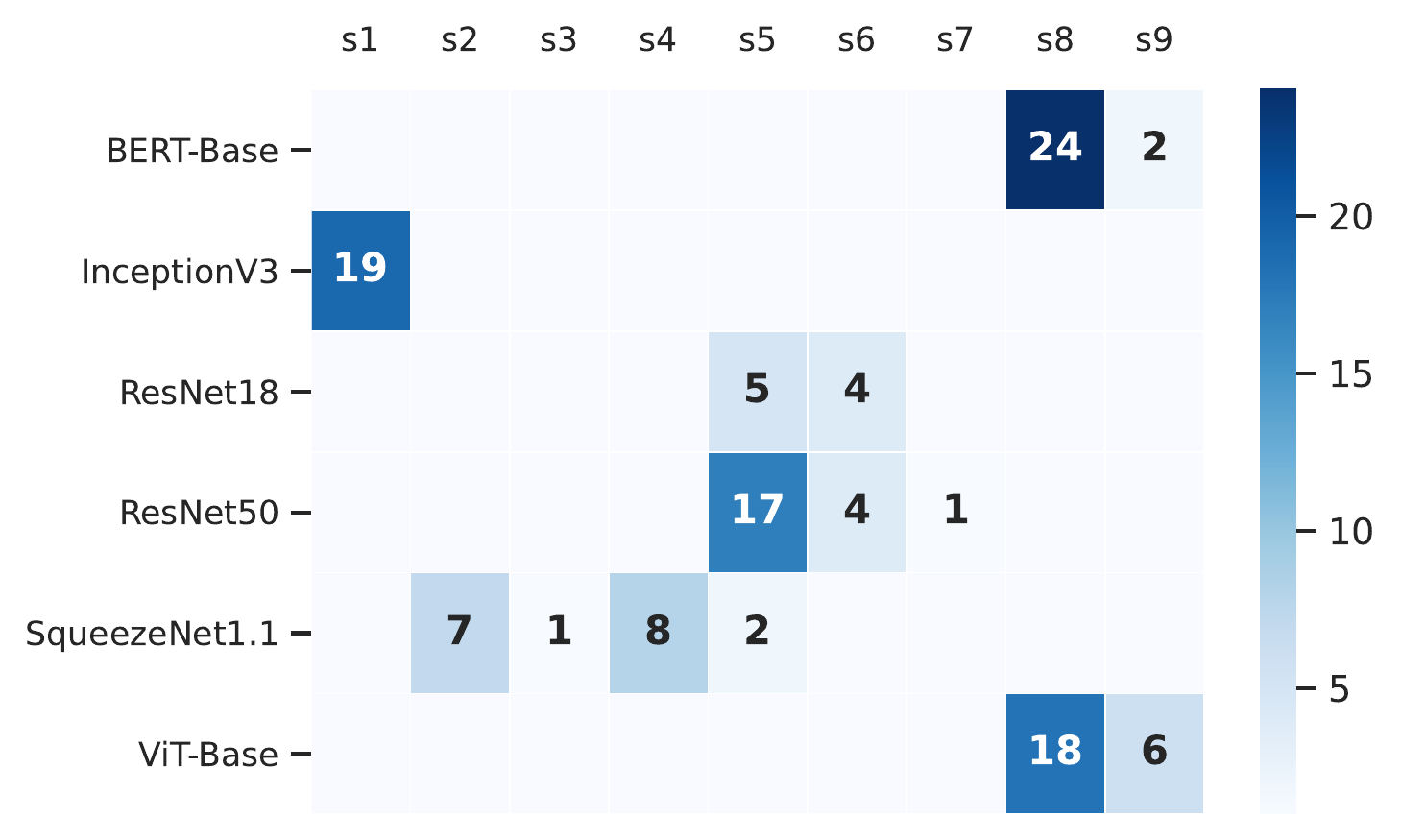}
  \caption[Heatmap of graph transformations applied by MB controller]{Heatmap showing the transformations applied by the trained controller acting inside the world model. Although there are over 100 possible transformations, we only show the transformations applied onto at least one graph. The counts for each transformation shows the number of times it has been applied. A higher count means that RLFlow was able to find a long sequence of transformations before termination. This can indicate RLFlow has discovered a performant graph via the transformation sequence.}
  \label{fig:eval:xfer-heatmap}
\end{figure}

Figure \ref{fig:eval:xfer-heatmap} shows a heatmap of the various graph transformations that have been applied by a trained model-based agent during evaluation. Notably, the optimisations applied to the ResNet18/50 graphs apply similar transformations, those targeting the convolutions in the network. The networks are composed of similar convolutional operators, albeit with different depths, and apply analogous transformations. For recurrent networks such as BERT and ViT, we perform relatively few transformations. This is in contrast to transformations performed by TASO that perform four substitutions, in comparison RLFlow that uses only two.

As we can infer from Figure \ref{fig:eval:xfer-heatmap}, and the results presented in TASO, there is a difference between the two approaches in both the specific transformations and the number of times each was applied. Despite this difference, we found that our method can achieve a similar level of performance in convolutional networks and outperform TASO when optimising transformer-based networks.  Primarily, this shows that there are many possible optimisation paths to achieve a performant, highly optimised network architecture.

\subsection{Effectiveness on Transformers}
Figure \ref{fig:eval:xfer-heatmap} shows that RLFlow applies two sub-graph substitutions to the BERT-Base and ViT-Base graphs. These two graphs use a transformer-style architecture that features many repeated layers of Multi-Head Attention, Addition and Normalisation. RLFlow automatically discovered and applied a sub-graph transformation that fused multiple element-wise additions into a single operation that provides a significant reduction in model runtime.

\begin{figure}[htb!]
  \centering
  \includegraphics[width=0.4\columnwidth]{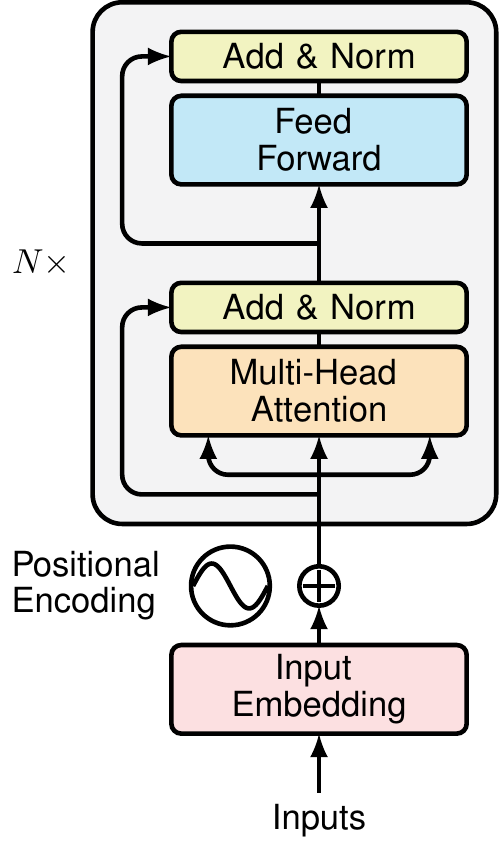}
  \caption{Transformer Encoder architecture. We only show the encoder slice of the whole transformer network. The architecture of ViT \cite{dosovitskiy2021image} is similar to the transformer architecture presented here, thus, RLFlow can apply the same optimisations that improve model runtime. Adapted from \cite{vaswani2017attention}.}
  \label{fig:eval:transformer}
\end{figure}

Figure \ref{fig:eval:transformer} shows a simplified version of Transformer encoder block used in state-of-the-art transformer networks. The network contains $N$ repeated blocks, in a similar fashion, the decoder uses repeated blocks of the Transformer Decoder. RLFlow can avoid greedily exploring the search space needlessly when the graph context doesn't change and instead fuses multiple addition/normalisation operations. In comparison, TASO fails to find the same sub-graph substitution sequence due to its cost-based search ignoring viable substitution sequences. Importantly, using this result, we note that it could be possible to directly implement such rules into the existing heuristics of TensorFlow and PyTorch to automatically improve the performance of transformer-style architectures.

\subsection{Summary}
In summary, the performance of a cost-directed search is dependent on the neural network architecture. RLFlow's approach performs better on specific architecture types (such as transformers), compared to convolutional networks (CNNs). The cost-directed search method might be altered and the appropriate optimisation methods for transforming the sub-graphs might be selected.

Recent work, PET \cite{wang2021pet}, has shown that CNNs can generally benefit more from partially equivalent transformations compared to the Transformer-based model. The CNN model provides more opportunities for transposing tensor shapes/layouts across dimensions, most of which are partially equivalent transformations, and as the CNNs access data in irregular ways, the approximate transformations are more applicable to achieve higher performance.

The RL agent can consider possible non-equivalent transformations during exploration, therefore, it could achieve better performance on CNN's. Such a modification only requires adding the partially equivalent transformations to the agent's action set. However, this is left for future work.

\section{Related Work}

\subsection{Optimisation of computation graphs}
To optimise computation graphs we employ a strategy by which we transform an input graph to alter its performance characteristics. Rule-based approaches such as those used in TensorFlow \cite{tensorflow2015whitepaper, 199317} and TVM \cite{chen2018tvm} use a pre-defined set of transformations that are applied greedily. In addition, recent work, such as \cite{jia2019optimizing,jia2019taso} automatically search for transformations to apply to the input graph with the modification that we allow performance decreasing transformations. Our work is similar as we use the same automated approach to automatically discover and verify the operator transformations in an offline manner, before the optimisation of the models.

\subsection{Model-based Reinforcement Learning}
Model-based RL is a class of reinforcement learning algorithms in which we aim to learn a model (or use a given model) of the real environment where an agent acts. The work in \cite{ha2018worldmodels} proposed a novel approach to learn a ``world model'' using recurrent neural networks; we take inspiration from such work and use world models and a policy optimisation algorithm as the controller in the world model. In contrast, alternative approaches have been proposed such as imagination-augmented agents \cite{weber2018imaginationaugmented} and model-based value estimation for model-free agents \cite{feinberg2018modelbased}. Furthermore, Nagabandi et al. \cite{nagabandi2017neural} proposed a method to combine the sample efficiency of model-based RL and the major benefit of model-free RL, stable performant agents. Our work differs as we use an RNN-based world model to simulate the environment dynamics. Other works such as \cite{robine2021smaller, hafner2021mastering} build discrete world models and train directly in latent space. Prior work on world models used a variation on a variational auto-encoders \cite{ha2018worldmodels,hafner2020dream} to generate a latent state of the pixel input, instead, we use a graph neural network \cite{battaglia2018relational} to generate a latent representation of the input computation graphs.

\subsection{Transformer Networks}
Transformer networks are a recent innovation, first described in its popular form by Vaswani et al. \cite{vaswani2017attention}. They described a Transformer encoder architecture to solve the Sequence-to-Sequence problem by handling long-range dependencies. Recently, Dosovitskiy et al. \cite{dosovitskiy2021image} described an approach to encode an RGB image into small patches with positional embedding for use in a Transformer encoder. RLFlow performs well on Transformer-style networks as they share a common architecture. A Transformer encoder is made of repeated blocks of addition, normalisation and multi-head attention operations for which RLFlow can discover and apply sub-graph transformations that results in a high throughput model.

\subsection{RL in Computer Systems}
Reinforcement Learning in Computer Systems is a relatively recent topic of research. In recent years there has been an increased focus on using model-free RL in a variety of systems environments. For example, in \cite{mirhoseini2017device, mirhoseini2018hierarchical, addanki2019placeto, paliwal2020reinforced}, reinforcement learning is used to optimise the placement of machine learning models to improve throughput. In \cite{app10196685}, model-based RL is used successfully to optimise the selection of bitrate when streaming data across a network. Although prior work has used a cost-directed search or manually defined heuristics, instead, we used model-based RL techniques to optimise the architecture of deep learning models.


\section{Conclusion}
In this work, we have shown how to apply model-based reinforcement learning techniques to optimise the architecture of neural networks. Our approach uses RL agents to select optimal actions that apply sub-graph transformations to computation graphs to improve on-device runtime. We performed experiments that show RL agents decreased the runtime on all of the six test graphs, each of which had unique properties and architectures. Notably, we provide evidence to support our claim that it is possible to learn a world model of the environment which is sufficiently accurate to enable the end-to-end training of an agent inside a fully imagined world model.

This work has highlighted that the performance of model-based agents trained inside a hallucinogenic is highly dependent on the accuracy of the world model. Inaccuracies in the model, can lead to compounding errors and thus the agent choosing sub-optimal, or invalid, actions that diverge the imagined state from the true environment state. Hence, there are still significant fundamental difficulties in training stable, accurate world models that can simulate the true environment; if one can train such a model by carefully tuning hyperparameters, we can gain substantial benefits through increased sample efficiency and decreased training time.

Finally, we also highlight that RLFlow performs notably well on Transformer-style architectures compared to the state-of-the-art. RLFlow has shown that it is possible to extract further performance from these models through a sequence of sub-graph transformations that combine operations in transformer-encoder blocks---providing up to a 5\% decrease in model runtime.


\bibliographystyle{ACM-Reference-Format}
\bibliography{ref}

\end{document}